%% file: top.tex
\pdfoutput=1
\documentclass[10pt,twocolumn,letterpaper]{article}

\usepackage{floatrow}
\usepackage{iccv}
\usepackage{times}
\usepackage{epsfig}
\usepackage{graphicx}
\usepackage{amsmath}
\usepackage{amssymb}
\usepackage{bm}
\usepackage{bbm}
\usepackage{algorithm}
\usepackage{algorithmic}
\usepackage[nopar]{lipsum}

\newfloatcommand{capbtabbox}{table}[][\FBwidth]
\usepackage{array}
\newcolumntype{?}{!{\vrule width 1pt}}
\newcolumntype{C}[1]{>{\centering\arraybackslash\hspace{0pt}}p{#1}}

\input{defs}

\DeclareMathOperator*{\argmax}{\arg\!\max}

\iccvfinalcopy 

\def\iccvPaperID{633} 

\begin{document}

\title{Deep Occlusion Reasoning \\ for Multi-Camera Multi-Target Detection}

\author{Pierre Baqu{\'{e}}$^1$ \quad\quad Fran{\c{c}}ois Fleuret$^{1,2}$ \quad\quad Pascal Fua$^1$ \\
$^1$CVLab, EPFL, Lausanne, Switzerland\\
$^2$IDIAP, Martigny, Switzerland\\
{\tt\small \{firstname.lastname\}@epfl.ch}
}

\maketitle

\begin{abstract}

People detection in single 2D images has improved greatly in recent years. However, comparatively little of this progress has percolated into multi-camera multi-people tracking algorithms, whose performance still degrades severely when scenes become very crowded. In this work, we introduce a new architecture that combines Convolutional Neural Nets and Conditional Random Fields to explicitly model those ambiguities. One of its key ingredients are high-order CRF terms that model potential occlusions and give our approach its robustness even when many people are present. Our model is trained end-to-end and we show that it outperforms several state-of-art algorithms on challenging scenes. 

\end{abstract}

\vspace{-0.2cm}

\input{intro}

\input{related}
\input{method}
\input{inference}
\input{training}
\input{implementation}
\input{evaluation}

\input{conclusion}

\clearpage

{\small
  \bibliographystyle{ieee}
  \bibliography{string,vision,learning,optim}
}

\end{document}

%% file: defs.tex
\newcounter{nbdrafts}
\makeatletter
\newcommand{\checknbdrafts}{
\ifnum \thenbdrafts > 0
\@latex@warning@no@line{**********************************************************************}
\@latex@warning@no@line{* The document contains \thenbdrafts \space draft note(s)}
\@latex@warning@no@line{**********************************************************************}
\fi}

\makeatother

\newcommand{\comment}[1]{}
\newcommand{\OURSWFT}[0]{{\bf Ours}}
\newcommand{\OURSNFT}[0]{{\bf Ours-No-FT}}
\newcommand{\OURSSHO}[0]{{\bf Ours-Simple-HO}}
\newcommand{\OURSNHO}[0]{{\bf Ours-No-HO}}
\newcommand{\OURSUSP}[0]{{\bf Ours-Unsuperv}}

 \newcommand{\POM}[0] {{\bf POM-CNN}}
 \newcommand{\CNN}[0] {{\bf RCNN-2D/3D}}

\newcommand{\EPFL}[0]{{\bf  EPFL}}
\newcommand{\ETHZ}[0]{{\bf  ETHZ}}
\newcommand{\PETS}[0]{{\bf  PETS}}

\newcommand{\vx}[0]{\vec{x}}
\newcommand{\vX}[0]{\vec{X}}

\newcommand{\mC}[0]{\mathcal{C}}

\newcommand{\mF}[0]{\mathcal{F}}

\newcommand{\mN}[0]{\mathcal{N}}
\newcommand{\mZ}[0]{\mathcal{Z}}
\newcommand{\mP}[0]{\mathcal{P}}

\newcommand{\KL}[0]{\text{KL}}

\newif\ifdraft
\drafttrue
\draftfalse

\ifdraft
 \newcommand{\PF}[1]{\textcolor{blue}{{\bf PF: #1}}}
 \newcommand{\pf}[1]{\textcolor{blue}{#1}}
 \newcommand{\PB}[1]{\textcolor{red}{{\bf PB: #1}}}
 \newcommand{\pb}[1]{\textcolor{red}{#1}}
  
 \newcommand{\FF}[1]{\textcolor{red}{{\bf FF: #1}}}
 
\else
 \newcommand{\PF}[1]{}
 \newcommand{\FF}[1]{}
 \newcommand{\PB}[1]{}
 \newcommand{\pf}[1]{ #1 }
 
 \newcommand{\pb}[1]{ #1 }
 
\fi

\def\mathclap#1{\text{\hbox to 0pt{\hss$\mathsurround=0pt#1$\hss}}}

\newcommand\blfootnote[1]{%
  \begingroup
  \renewcommand\thefootnote{}\footnote{#1}%
  \addtocounter{footnote}{-1}%
  \endgroup
}

%% file: intro.tex

\section{Introduction}

Multi-Camera Multi-Target Tracking (MCMT) algorithms have long been effective at tracking people in complex environments. Before the emergence of Deep Learning, some of the most effective methods relied on simple background subtraction, geometric and sparsity constraints, and occlusion reasoning~\cite{Fleuret08a,Berclaz11,Alahi11}.  Given the limited discriminative power of background subtraction, they work surprisingly well as long as there are not too many people in the scene. However, their performance degrades as people density increases, making the background subtraction used as input less and less informative. 

\input{teaser}

Since then, Deep Learning based people detection algorithms in single images~\cite{Ren15,Liu16,Zhang16a} have become among the most effective~\cite{Zhang16a}. However, their power has only rarely been leveraged for MCMT purposes. Some recent algorithms, such as the one of~\cite{Xu16}, attempt to do so by first detecting people in single images, projecting the detections into a common reference-frame, and finally putting them into correspondence to achieve 3D localization and eliminate false positives. As shown in Fig.~\ref{fig:teaser}, this is prone to errors for two reasons. First, projection in the reference frame is inaccurate, especially when the 2D detector has not been specifically trained for that purpose. Second, the projection is usually preceded by Non Maximum Suppression (NMS) on the output of the 2D detector, which does not take into account the multi-camera geometry to resolve ambiguities.

Ideally, the power of Deep Learning should be combined with occlusion reasoning much earlier in the detection process than is normally done.  To this end, we designed a joint CNN/CRF model whose posterior distribution can be approximated by Mean-Field inference using standard differentiable operations. Our model is trainable end-to-end and can be used in both supervised and unsupervised scenarios. 


More specifically, we reason on a discretized ground plane in which detections are represented by boolean variables. The CRF is defined as a sum of innovative high-order terms whose values are computed by measuring the discrepancy between the predictions of a generative model that accounts for occlusions and those of a CNN that can infer that certain  image patches look like specific body parts.
To these terms, we add unary and pairwise ones to increase robustness and model physical repulsion constraints. 


To summarize, our contribution is a joint CNN/CRF pipeline that performs detection for MCMT purposes in such a way that NMS is not required. Because it explicitly models occlusions, our algorithm operates robustly even in crowded scenes. Furthermore, it outputs probabilities of presence on the ground plane, as opposed to binary detections, which can then be linked into full trajectories using a simple flow-based approach~\cite{Berclaz11}.

%% file: teaser.tex
\begin{figure*}[ht!]
\begin{center}
\begin{tabular}{ccc}
\includegraphics[width=0.31\textwidth]{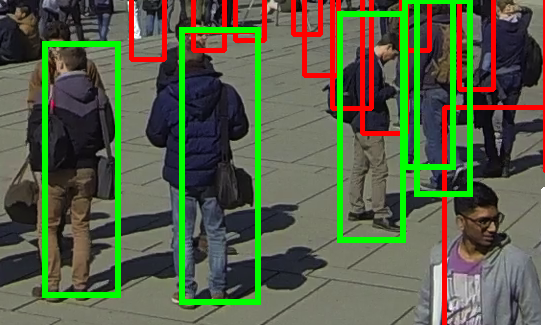}  &
\includegraphics[width=0.31\textwidth]{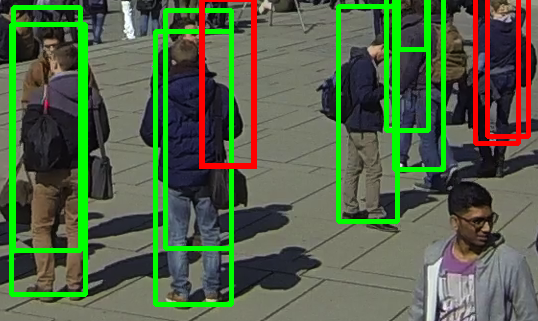}   &
\includegraphics[width=0.31\textwidth]{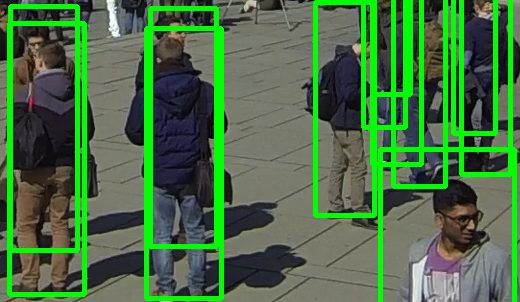}   \\[-0.1cm]
\CNN{} & \POM{} & {\bf Ours}
\end{tabular}
\end{center}
\vspace{-0.3cm}
   \caption{Multi-camera detection in a crowded scene. Even though there are 7 cameras with overlapping fields of view, baselines inspired by earlier approaches----\CNN{} by~\cite{Xu16} and \POM{} by~\cite{Fleuret08a}, as described in Section~\ref{sec:results}---both generate false positives denoted by red rectangles and miss or misplace a number of people, whereas ours does not. This example is representative of the algorithm's behavior and is best viewed in color. Please see supplementary material for results on a video sequence. }
\label{fig:teaser}
\vspace{-0.3cm}
\end{figure*}

%% file: related.tex
\section{Related Work}
\label{sec:related}

In this section, we first discuss briefly recent Deep Learning approaches  to people detection in single images. We then move on to multi-image algorithms  and techniques for combining CNNs and CRFs. 

\subsection{Deep Monocular Detection}
\label{related:subsec:monocular}

As in many  other domains,  CNN-based algorithms~\cite{Ren15,Liu16,Redmon16} have become for very good for people detection in single images and achieve state-of-the-art performance~\cite{Zhang16a}. Algorithms in this class usually first propose potential candidate bounding boxes with scores assigned to them. They then perform Non-Maximum Suppression (NMS) and return a final set of candidates. The very popular method of~\cite{Ren15} performs both steps in a single CNN pass through the image. It returns a feature map in which a feature vector of constant dimension is associated to each image pixel. For any 2D bounding-box of any size in that image, a feature vector of any arbitrary dimension can then be computed using Region Of Interest (ROI) pooling and fed to a classifier to assess whether the bounding box does indeed correspond to a true detection.

While this algorithm has demonstrated its worth on many benchmarks, it can fail in crowded scenes such as the one of  Fig.~\ref{fig:teaser}. This is perennial problem of single-image detectors when people occlude each other severely. One solution to this problem is to rely on cameras with overlapping fields of view, as  discussed below.

\subsection{Multi-Camera Pedestrian Detection}
\label{related:subsec:multicam}

Here, we distinguish between recent algorithms that rely on Deep Learning but do not explicitly account for occlusions  and older ones that model occlusions and geometry but appeared before the Deep Learning became popular. Our approach can be understood as a way to bring together their respective strengths.

The recent algorithm of~\cite{Xu16} runs a monocular detector similar to the one of~\cite{Ren15} on multiple views and infers people ground locations from the resulting detections. However, this method is prone to errors both because the 2D detections are performed independently of each other and because combining them by projecting them onto the ground plane involves reprojection errors and ignores occlusions. Yet, it is representative of the current MCMT state-of-the-art and is benchmarked against much older algorithms~\cite{Fleuret08a,Berclaz11} that rely on background subtraction instead of a Deep Learning approach. 

These older algorithms use multiple cameras with overlapping fields of view to leverage geometrical or appearance consistency across views to resolve the ambiguities that arise in crowded scenes and obtain accurate 3D localisation~\cite{Fleuret08a,Alahi11,Peng15b}.  They rely on Bayesian inference and graphical models to enforce detection sparsity.  For example, the Probabilisitic Occupancy Map (POM) approach~\cite{Fleuret08a}  takes background subtraction images as input  and relies on Mean Field inference to compute probabilities of presence in the ground plane.  More specifically, given several cameras with overlapping fields of view of a discretized ground plane, POM first performs background  subtraction.  It then uses a generative model that represents humans as simple rectangles in order to create synthetic ideal images that would be observed if people were
at given locations. Under this model of the image given the true occupancy, it approximates the probabilities of occupancy at every location using Mean Field inference. Because the generative model explicitly accounts for occlusions, POM is robust and often performs well. But it relies on background subtraction results as its only input, which is not discriminative enough when the people density increases, as shown in Fig.~\ref{fig:teaser}. The algorithm of~\cite{Alahi11}  operates on similar principles as POM but introduces more sophisticated human templates. Since it also relies on background subtraction, it is subject to the same limitations when the people density increases. And so is the algorithm of~\cite{Peng15b}  that introduces a more complex Bayesian model to enhance the results of~\cite{Alahi11}. 

\comment{Recently,~\cite{Chavdarova17} classified 3D detections proposals using an approach similar the the ones proposed in~\ref{related:subsec:monocular} except that features are extracted simultaneously from multiple views. However, they still use a greedy non-maximum suppression step and don't reason jointly about final detections and occlusions. Furthermore, their algorithm produces mere detections and not a richer probabilistic occupancy map. \PF{\cite{Chavdarova17}  is still an ArXiv paper that we may or may not talk about to save space. And if we talk about it, we have to compare. }}

\subsection{Combining CNNs and CRFs}

Using a CNN to compute potentials for a Conditional Random Field (CRF) and training them jointly for structured prediction purposes has received much attention in recent years~\cite{Lecun06,Do10,Domke13,Zheng15,Arnab15,Kirillov15b,Larsson17,Bagautdinov17}. However, properly training the CRFs remains difficult because many interesting models yield intractable inference problems.  A popular workaround is to optimize the CRF potentials so as to minimize a loss  defined on the output of an inference algorithm.  Back Mean-Field~\cite{Domke13,Zheng15,Arnab15,Larsson17} has emerged as a promising  way to do this. It relies on the fact that the updates steps during Mean-Field inference are continuous and parallelizable~\cite{Baque16}. It is therefore possible to represent these operations as additional layers in a Neural Network and back-propagate through it. So far, this method has mostly been demonstrated either for toy problems or for semantic segmentation with attractive potentials, whereas our approach also requires repulsive potentials. 


%% file: method.tex

\section{Modeling Occlusions in a CNN Framework}

The core motivation behind our approach is to properly handle occlusions, while still leveraging the power of CNNs. To do so, we must model the interactions between multiple people who occlude each other but may not be physically close to each other. Our solution is to introduce an {\it observation space}; a generative model for observations given where people are located in the ground plane; and a discriminative model that predicts expected observations from the images.  We then define a loss function that measures how different the CNN predictions are from those generated by the model. Finally,  we use a  Mean-Field approach with respect to probabilities of presence in the ground plane to minimize this loss. We cast this computation in terms of minimizing the energy of a Conditional Random Field in which the interactions between nodes are non-local because the people who occlude each other may not be physically close, which requires long range high-order terms. 

In the remainder of this section, we first introduce the required notations to formalize our model.  We then define a CRF that only involves high-order interaction potentials. Finally, we describe a more complete one that also relies on unary and pairwise terms. 

\subsection{Notations}
\label{sec:notations}

We discretrize the ground plane in grid cells and introduce Boolean variables that denote the presence or absence of someone in the cell. 
Let us therefore consider a discretized ground plane containing $N$ locations.  Let $Z_i$ be the boolean variable that denotes the presence of someone at location $i$. Let us assume we are given  $C$ RGB images $I^c$ of size $H^c \times W^c$ from multiple views $1\leq c \leq C$ and $I = \{I^1,\dots,I^C\}$. For each ground plane location $i$ and camera $c$, let the smallest rectangular zone  containing the 2D projection of a human-sized 3D cylinder located at $i$ be defined by its top-left and bottom-right coordinates $T^c_i$ and $B^c_i$. For a pixel $k \in \{1,\dots,H^c \} \times \{1,\dots,W^c \}$, let $L^c_k$ be the set of such projections that contain $k$. 


We also introduce a CNN that defines an operator $\mF(\cdot ; \theta_{F})$, which takes as input the RGB image of camera $c$ and outputs a feature map $\mF^c = \mF(I^c; \theta_{F})$, where $\theta_{F}$ denotes the network's parameters. It contains
a $d$-dimensional vector  $\mF^c_k$ for each pixel $k$.

\subsection{High-Order CRF}
\label{sec:hoCRF}
\input{ho}

We take the energy of our CRF to be a sum of High-Order potentials $\psi^{c,k}_\text{h}$, one for each pixel. They handle jointly detection, and occlusion reasoning while removing the need for Non-Maximum Suppression. Each of these potentials use Probability Product Kernels~\cite{Jebara04} to represent the agreement between a generative model and a discriminative model over the  {\it observation space}, at a given pixel, as depicted in Fig.~\ref{fig:ho}. We therefore write 
\begin{eqnarray}
P(Z;I) & = & \dfrac{1}{\mZ} \exp \psi_\text{h}(Z;\mF(I;\theta_F))\;, \label{eq:ho_crf} \\
\psi_\text{h}(Z;\mF) &=& \hspace{-0.3cm}\sum_{ 1\leq c \leq C, k \in  \{1,\dots,H^c \} \times \{1,\dots,W^c \} } \psi^{c,k}_\text{h}(Z; \mF^c_k) \; . \nonumber
\end{eqnarray}
Assuming we know the values of the occupancy variables $Z$, the generative model computes distributions over the set of {\it observations}. 
For each pixel  in each image, it considers the corresponding line of sight and computes a distribution of vectors depending on the probability that it actually belongs to the successive people it traverses. This results in images whose pixels are vectors representing a distribution of 2D vectors, the  observations, as depicted in the top row of Fig~\ref{fig:ho}. Our discriminative model relies on a CNN which tries to predict similar distributions of 2D vectors, directly by looking at the image. 
For ease of understanding,  we first present in more details a simple version of our High-Order potentials $\psi^{c,k}_\text{h}$. It assumes that our observations are zeros and ones at every pixel. The discriminative model therefore acts much as the background subtraction algorithms used in~\cite{Fleuret08a} did.  We then extend them to take into account the 2D vector output of our discriminative model.

\subsubsection{Simple Generative Model} 
\label{sec:simpleGenerative}
We first introduce a  binary {\it observation} variable $X^c_k \in \{0,1\}$ over which we define two distributions $P^g$ and $P^d$  produced by the generative and discriminative model respectively. We take the  distribution $P^g$ to be
\begin{align}
P^g(X^c_k =1  | Z)= 0, & \mbox{ if }  Z_i = 0 \; \forall i \in  L^c_k \label{eq:method:pom} \;,\\
P^g(X^c_k =1  | Z) = 1 &\mbox{ otherwise,}\nonumber
\end{align}
and the discriminative one $P^d$ to be $P^d(X^c_k | \mF^c_k) = f_\text{b}(\mF^c_k;\theta_{b})$, where $F^c_k$ is the $d$-dimensional feature vector  associated to pixel $k$ introduced above and $f_\text{b}$ is a Multi-Layer Perceptron (MLP) with weights $\theta_{b}$. In other words, $ f_\text{b}$ plays the role of a CNN-based semantic segmentor or background-subtraction. 

For each pixel, we then take the high-order potential to be the dot product between the distributions
\begin{small}
\begin{equation}
\label{method:ho_general}
\psi^{c,k}_\text{h}(Z;\mF^c_k) = \mu_\text{h} \log \int \limits_{\mathclap{X^c_k \in \{0,1 \}} } P^g(X^c_k | \{Z_i\}_{i \in L^c_k}) P^d(X^c_k | \mF^c_k) \;,
\end{equation}
\end{small}
as in the probability product kernel method of~\cite{Jebara04}. Intuitively, $\psi^{c,k}_\text{h}$ is high when the segmentation produced by the network matches the projection of the detections in each camera plane using the simple generative model of Eq.~\ref{eq:method:pom}. \pb{$\mu_\text{h}$ is an energy scaling parameter.}

\subsubsection{Full Generative Model} 
\label{sec:fullGenerative}

The above model correctly accounts for occlusions and geometry but ignores much image information by focusing on background / foreground decisions. To refine it, we model the part of the bounding-box a pixel belongs to rather than just the fact that it belongs to a bounding-box. To this end, we redefine the Boolean auxiliary variable $X^c_k$ as
\begin{equation}
\vX^c_k \in \{0\} \cup \mathbb{R}^2 \;,
\end{equation}
where the label $0$ represents background as before, and a label in $\mathbb{R}^2$ denotes the displacement with respect to the center of the body of the visible person at this pixel location. 

To extend the simple model and account for what part of a bounding-box pixel $k$ belongs to if it does,  we sample from the distribution $P^g(\vX^c_k |  \{Z_i\}_{i \in L^c_k} )$. To this end, let us assume without loss of generality that the  $L^c_k$ are ordered by increasing distance to the camera, as shown in the top left corner of Fig.~\ref{fig:ho}. We initialize the variables $\vX^c_k$ to 0. Then, for each $i$ in $L^c_k$ such that $Z_i = 1$, we draw a boolean random variable $O_i$ with  fixed expectancy $o$. If $O_i = 1 $, then
\begin{small}
\begin{eqnarray}
\vX_k & = & \vx^i_k \;, \\
          & = & \left ( \dfrac{k_x - 0.5({T^c_i}_x+{B^c_i}_x)}{{B^c_i}_x - {T^c_i}_x} ; \dfrac{k_y - 0.5({T^c_i}_y+{B^c_i}_y)}{{B^c_i}_y - {T^c_i}_y} \right) \;, \nonumber
\end{eqnarray}
\end{small}
that is, the relative location of pixel $k$ with respect to the projection of detection $i$ in camera $c$, as depicted in the upper right corner of  Fig.~\ref{fig:ho}.

We define the distribution $P^d(\vX_k | \mF^c_k)$ as an $M$-Modal Gaussian Mixture
\begin{small}
\begin{align}
\label{model:gaussian_mixture}
P^d(\vX_k = 0) &= f_\text{b}(\mF^c_k;\theta_{b}) \;, \\ 
P^d(\vX_k |\vX_k \neq 0  ) &= \sum \limits_{ 1 \leq m \leq M} f_\text{h}(\mF^c_k;\theta_\text{h})_m \mN(\vX_k - \alpha_m ; \sigma_m) \; , \nonumber
\end{align}
\end{small}
as depicted in the bottom right corner of Fig.~\ref{fig:ho}. As a result, $P^d(\vX_k = 0)$ is the same as in the simple model but $P^d(\vX_k |\vX_k \neq 0 )$ encodes more information.
$(\alpha_m, \sigma_m)$ are Gaussian parameters learned for each mode $m$. 
$f_\text{h}$ is a MLP parametrized by $\theta_\text{h}$ that outputs $M$ normalized real probabilities where M is a meta-parameter of our model. Similarly, $f_\text{b}(\mF^c_k;\theta_{b})$ is a background probability.

Finally, as in Eq.~\ref{method:ho_general}, we take our complete potential to be
\begin{small}
\begin{equation}
\label{method:ho_specific}
\hspace{-0.2cm} \psi^{c,k}_\text{h}(Z;\mF^c_k) = \mu_\text{h}  \log \int \limits_{\mathclap{\vX_k \in \{0\} \cup \mathbb{R}^2 }} P^g(\vX_k | \{Z_i\}_{i \in L^c_k}) P^d(\vX_k | \mF^c_k)   \;.
\end{equation}
\end{small}

\subsection{Complete CRF}
\label{sec:completeCRF}

To increase the robustness of our CRF, we have found it effective to add, to the high-order potentials of  Eq.~\ref{eq:ho_crf}, unary and pairwise ones to exploit additional image information. We therefore write our complete CRF model as
\begin{small}
\begin{eqnarray}
\label{method:completeCRF}
P(Z;I)        & \hspace{-0.2cm} = & \hspace{-0.2cm} \dfrac{1}{\mZ} \exp \psi(Z;\mF)\;, \label{eq:crf} \\
\psi(Z;\mF) &\hspace{-0.2cm} =& \hspace{-0.2cm} \psi_\text{h}(Z;\mF)  + \sum_{i \leq N} \psi^i_\text{u}(Z_i ; \mF)  + \sum_{i \leq N, j \leq N} \psi_\text{p}(Z_i,Z_j) , \nonumber
\end{eqnarray}
\end{small}
where $\psi_\text{h}$ is the high-order CRF of Eq.~\ref{eq:ho_crf}, the $\psi^i_\text{u}$ are unary potentials, and $\psi_\text{p}$ pairwise ones, which we describe below. 




\subsubsection{Unaries}
\label{sec:unaries}
The purpose of our unary potentials is to provide a prior probability of presence at a given location on the ground, before 
considering the occlusion effect and non maximum suppression. 
For each location $i$ and camera $c$, we use a CNN  $f_\text{u}(T^c_i,B^c_i,\mF^c)$, which outputs a probability of presence of a person at location $i$. $f_\text{u}$ works by extracting a fixed size feature vector from the rectangular region defined by $T^c_i,B^c_i$ in $\mF^c$, using an ROI pooling layer~\cite{Ren15}. A detection probability is finally estimated using an MLP. Estimates from the multiple cameras are pooled through a $max$ operation
 \begin{equation}
 \label{method:unary}
\psi^i_\text{u}(Z_i ; \mF) = \mu_\text{u} Z_i \max_{c} \log \dfrac{f_\text{u}(T^c_i,B^c_i,\mF_c)}{1-f_\text{u}(T^c_i,B^c_i,\mF_c)} \;,
 \end{equation}
 where \pb{$\mu_\text{u}$ is a scalar that controls the importance of unary terms compared to others.}
 
\subsubsection{Pairwise}

The purpose of our pairwise potentials is to  represent the fact that two people are unlikely to stand too close to each others. 
\comment{
For all pairs of locations $(i,j)$, let $E^{i,j}_\text{p} = E_\text{p} \mathbbm{1}[|x_i - x_j|+|y_i - y_j| < r]$, where $E_\text{p}$ is a large positive constant and $r$ a predefined interaction radius. We write
}
For all pairs of locations $(i,j)$, let $E^{i,j}_\text{p} = E_\text{p}[|x_i - x_j|;|y_i - y_j| ]$, where $E_\text{p}$ is a 2D kernel function of of predefined size. We write
\begin{equation}
 \label{method:binary}
\psi_\text{p}(Z_i,Z_j) = - E^{i,j}_\text{p} Z_i Z_j 
\end{equation}
for locations that are closer to each other than a predefined distance and 0 otherwise. 


%% file: ho.tex
\begin{figure}[ht!]
\begin{center}
\includegraphics[width=1.0\textwidth]{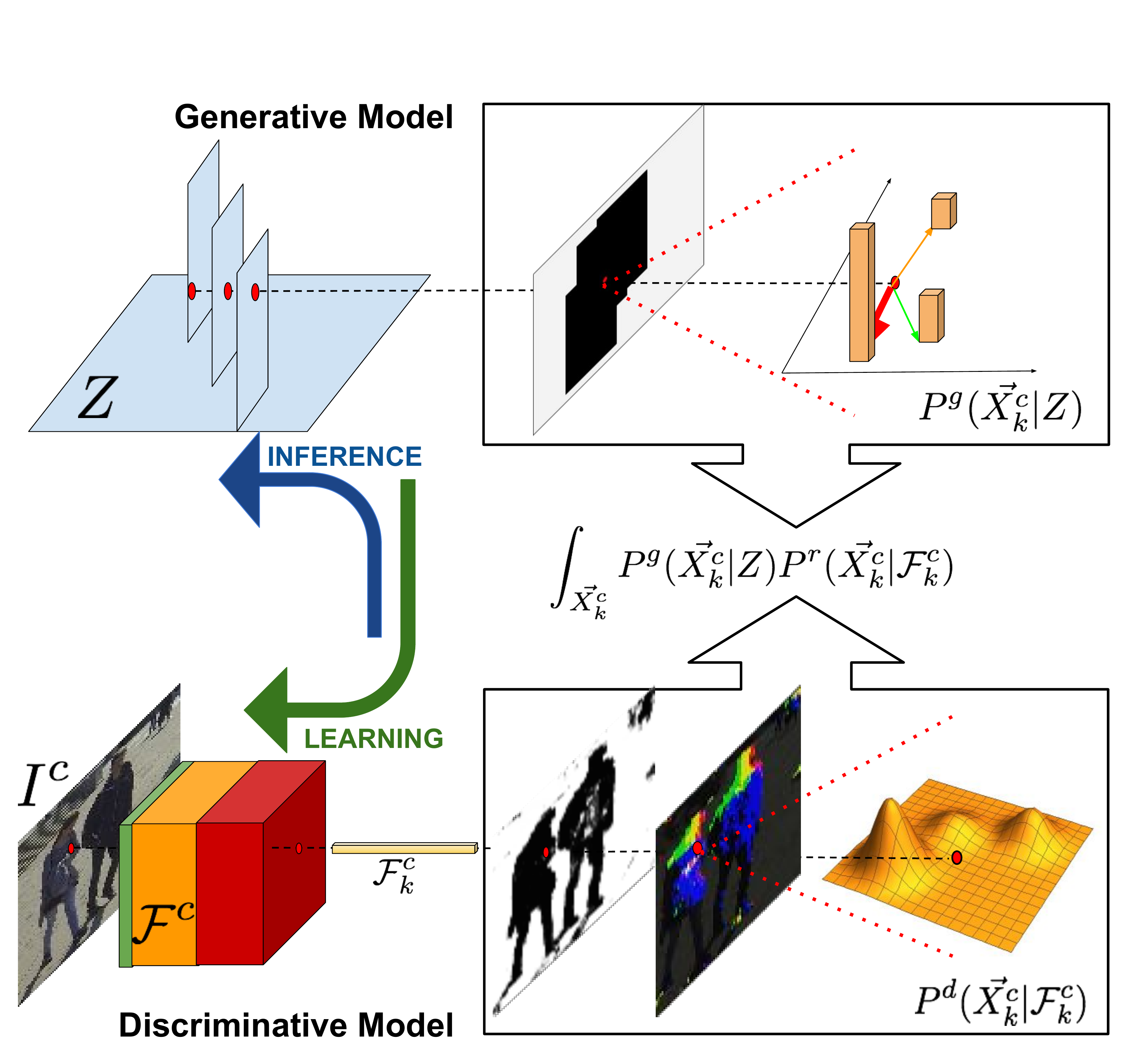} \\
\end{center}
\vspace{-0.2cm}
   \caption{Schematic representation of our High-Order potentials as described in Section~\ref{sec:fullGenerative}.}
   \label{fig:ho}
\vspace{-0.4cm}
\end{figure}
\comment{
    which links generative and discriminative models. For each pixel $k$, we introduce an auxiliary variable $\vX_k \in \{0\} \cup \mathbb{R}^2$ which represents Backgroud/Foreground and if foreground, the region of the body this pixel belongs to.
   Generative model to compute $P^g(\vX_k | \{Z_i\}_{i \in L^c_k})$: We consider all the bounding-boxes which intersect pixel $k$ and are present ($Z_i = 1$) ordered by closeness to the camera and use a probabilistic occlusion model to obtain a distribution over $ \vX_k$. 
   Discriminative model to compute $P^r(\vX_k | I^c)$: We use Fully-Convolutional Network to generate a distribution over
   $ \vX_k$ in the form of a Multi-Modal Gaussian Distribution. Finally, we use a the dot-product between $P^g$ and $P^r$ to represent the closeness of the two models. }

%% file: inference.tex

\section{Inference and Derivation}
\label{sec:inference}
Given the CRF of Eq.~\ref{eq:crf} and assuming all parameters known, finding out where people are in the ground plane amounts to minimizing $\psi$ with respect to $Z$, the vector of binary variables that indicates which ground locations contain someone, which amounts to computing a Maximum-a-Posteriori of the posterior $P$. Instead of doing so directly, which would be intractable, we use Mean-Field inference~\cite{Wainwright08}  to approximate of $P$ by a fully-factorised distribution $Q$. As in~\cite{Fleuret08a}, this produces a Probability Occupancy Map, that is, a probability of presence $Q(Z_i =1)$, at each location, such as the one depicted by Fig.~\ref{fig:pom}. 
\input{pom}

\comment{
We use the well known variational Mean-Fields in order to approximate the posterior $P(Z ; I)$ by a fully factorized distribution 
\begin{equation}
Q(Z) = \prod_i Q_i(Z_i) \;,
\end{equation}
which minmizes the KL-Divergence
\begin{equation}
\label{method:MF-objective}
\KL(Q ||P) = \sum_Z Q(Z)\log \left(\dfrac{Q(Z)}{P(Z)} \right) \;.
\end{equation}

In the context of pedestrian detection, we find three advantages to this method. First,  the output is a proper approximation of P --- as opposed to marginals or Maximum-a-Posteriori --- and is conveniently interpreted as a probabilistic occupancy map, which is very useful for tracking  and surveillance~\cite{Fleuret08a,Berclaz11,Berclaz08b}. Furthermore, we can compute efficiently the gradient steps to minimise the KL-Divergence, as we will see later. Finally, the inference iterates are fully differentiable, which makes it possible to use the Back Mean-Field method~\cite{Domke13} and train jointly the whole pipeline.

The recent work of~\cite{Baque16} proposed a natural-gradient descent scheme for optimizing the MF objective of Eq.~\ref{method:MF-objective}. They show that the gradient step to be taken at each iteration with respect to the natural parameters $\eta_i$ of the MF distribution $Q$, where $Q(Z_i) \propto \exp(Z_i \eta_i)$ , is proportional to

}
To perform this minimization, we rely on the natural-gradient descent scheme of~\cite{Baque16}. It involves taking gradient steps that are proportional to 
\begin{small}
\begin{equation}
\label{method:MF-update}
\hspace{-0.2cm}\nabla_{\eta_i} = \mathbb{E}_{Q} \left[ \left(\psi(Z,\mF) \right) | Z_i = 1 \right ] - \mathbb{E}_{Q} \left[  \left(\psi(Z,\mF) \right) | Z_i = 0 \right ] ,
\end{equation}
\end{small}
for each location $i$. The contribution to $\nabla_{\eta_i}$ of the unaries derives straightforwardly from Eq.~\ref{method:unary}.  Similarly, the one of the pairwise potentials of Eq.~\ref{method:binary} is
\begin{small}
\begin{align}
(\nabla_{\eta_i})_\text{p} &= -\sum_j E^{i,j}_\text{p} Q_j(Z_j = 1)\;,\label{method:pairwise-update}\\ 
& =  -\sum_j E_\text{p}[|x_i - x_j|,|y_i - y_j|] Q_j(Z_j = 1)\;, \nonumber
\end{align}
\end{small}
which can be implemented as a convolution over the current estimate of the probabilistic occupancy map $Q$ with the two dimensional kernel $E_\text{p}[.,.]$. This makes it easy to unroll the inference steps using a Deep-Learning framework.

Formulating the contributions of the higher-order terms of Eq.~\ref{method:ho_specific} is more involved and requires simplifications. We first approximate the Gaussians used in Eq.~\ref{model:gaussian_mixture} by a function whose value is $1$ in $B_m$ and $\epsilon$ elsewhere, where $B_m$ is the rectangle of center $\alpha_m$ and half-size $3 \sigma_m$. Note that this approximation is only used for inference purposes, and that during training, it keeps its original Gaussian form. We then threshold the Gaussian weights $f_\text{h}$ resulting in the binary approximation $\tilde f_\text{h}$. This yields a binary approximation $\widetilde R_\text{h}(\vX_k) $ of $R_\text{h}(\vX_k) $. Note that the corresponding approximate potential $ \widetilde \psi^{c,k}_\text{h}(Z,\mF^c_k) $ can be either $O(\log \epsilon)$, if $P(\vX_k,b_k = 1; Z) = 0 $ for all $\vX_k$ such that $R_\text{h}(\vX_k) > \epsilon $ or $O(\log(1))$. Hence, the configurations where  $\psi^{c,k}_\text{h}(Z,\mF^c_k) = O(\log \epsilon)$ will dominate the others when computing the expectancies. This yields the approximation of Eq.~\ref{method:MF-update},
\begin{small}
\begin{equation}
\widetilde {\nabla \eta_i} = -C ( \mathbb{E}_{Q} \left[ \Delta(Z) | Z_i = 1 \right ] - \mathbb{E}_{Q} \left[  \Delta(Z)  | Z_i = 0 \right ] ) \;,
\end{equation}
\end{small}
where $C = -log{\epsilon}$ is a constant and $\Delta(Z)$ is a binary random variable, which takes value 1 if  $\widetilde \psi^{c,k}_\text{h}(Z,\mF^c_k) = 0$, and 0 otherwise. 
\comment{
\begin{align}
1 &\mbox{ if }  \widetilde \psi^{c,k}_\text{h}(Z,\mF^c_k) = 0 \; , \\
0 &\mbox{ if }  \widetilde \psi^{c,k}_\text{h}(Z,\mF^c_k) >  0 \;. \nonumber
\end{align}
}
Note that $\psi^{c,k}_\text{h}(Z,\mF^c_k) = O(\log(1)) $ iff
\begin{small}
\begin{equation}
\exists i \leq N, m \leq M  \mbox{ s.t } \widetilde f_\text{h}(\mF^c_k;\theta_\text{h})_m = 1 \text{ and } \vX^i_k \in B_m \;.
\end{equation}
\end{small}
\comment{
\begin{align}
\exists i \in \{1,\dots,N\}, 1\leq m \leq M  \text{ s.t } \\
\widetilde f_\text{h}(\mF^c_k;\theta_\text{h})_m = 1 \text{ and } \vX^i_k \in B_m .
\end{align}
}
\comment{
\begin{equation}
 B_m  = ([(\alpha_m)_x - 3 \sigma_x,(\alpha_m)_x + 3 \sigma_x ];[(\alpha_m)_y - 3 \sigma_y,(\alpha_m)_y + 3 \sigma_y ]) 
\end{equation}
}
This means that for each pixel $k$, given a thresholded output from the network $\widetilde f_\text{h}(\mF^c_k;\theta_\text{h})$, we obtain a list of {\it compatible explanations} $\mC_k \subset \{1,\dots,N\}$ such that pixel $k$ defines a very simple pattern-based potential of the form $1$ if  $Z_i = 0 \; \forall i \in  \mC_k$, $0$ otherwise, which is similar to the potentials used in the Mean-Fields algorithms of~\cite{Vineet14,Fleuret08a,Kohli12,Arnab15,Baque17}. In the supplementary material, we see how this operation can be implemented efficiently using common Deep-Learning operations and integral-images.


\comment{
\begin{figure}[ht!]
\begin{center}
\begin{tabular}{@{}cc}
\includegraphics[width=0.248\textwidth]{illustrations/method_gaussians.png} &
\includegraphics[width=0.42\textwidth]{{illustrations/method_parts}.png}  \\
(a) & (b) \\
\end{tabular}
\end{center}
\vspace{-0.2cm}
   \caption{For, Gaussian Mixture Elements $m = 1, \dots, 3$  - - (a) Rectangles $B_m$ representing learned body parts.(b) Colored representation of $f_\text{h}(\mF^c_k;\theta_\text{h})_m$ for the pixels $k$ in the image.}
\label{fig:method:gaussians}
\vspace{-0.3cm}
\end{figure}
}


%% file: pom.tex

\begin{figure}[ht!]
\makebox[\textwidth][c]{
\begin{tabular}{cc}
\includegraphics[width=0.46\textwidth]{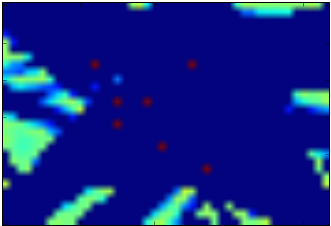}  &
\includegraphics[width=0.54\textwidth]{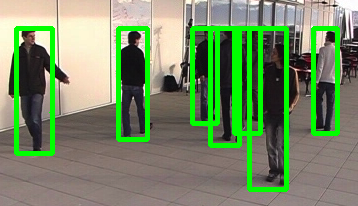} 
 \\[-0.1cm]
(a) & (b)
\end{tabular}
}
\vspace{0.1cm}

\caption{Output. (a) Given a set of images of the same scene, ours algorithm produces a Probabilistic Occupancy Map, that is, a probability of presence at each location of the ground plane. Red values indicate probabilities close to 1 and blue ones values close to zero. (b) Because the probabilities are very peaked, they can easily be thresholded to produce detections whose projections are the green boxes in the original image(s).}
\label{fig:pom}
\vspace{-0.3cm}
\end{figure}

%% file: training.tex

\section{Training}
\label{section:training}

We now show how our model can be trained first in a supervised manner and then in an unsupervised one. 

\subsection{Supervised Training}
\label{method:training:supervised}

Let us first assume that we observe $D$ data point $(Z^0,I^0),\dots,(Z^D,I^D)$, where $I^d$ represents a multi-view image and $Z^d$ the corresponding ground truth presences. The purpose of training is then to optimize the network parameters $\theta_F, \theta_\text{u}, \theta_\text{h}$ defined in Sections~\ref{sec:notations},~\ref{sec:unaries} and~\ref{sec:fullGenerative} respectively, the gaussian parameters $\alpha, \sigma$ of Eq.~\ref{model:gaussian_mixture} and the energy-scaling meta-parameters $\mu_\text{u},\mu_\text{h}$ of Eqs.~\ref{method:unary} and~\ref{method:ho_general} to maximize $\sum \limits_{d \leq D} \log P(Z^d;I^d)$. It cannot be done directly using Eq.~\ref{eq:crf} because computing the partition function $\mZ$ is intractable. 

\comment{
Intuitively, if one ignored the normaliser term when optimising the parameters, two things would happen. The potentials $\psi$ would not be prevented from growing infinitely for all configurations and would converge toward $+\infty$ without modelling the fact that some choices for $Z$ are more likely than the others. Furthermore, the combined effect of the potentials on the distribution would be ignored and nothing guarantees that potentials which make sense separately would still do so when combined into one function. 
}

\paragraph{Back Mean-Field} 

An increasingly popular work-around is to optimize the above-mentioned parameters to ensure that the output of the Mean-Field inference fits the ground truth. In other terms, let $Q_{\theta_F, \theta_\text{u}, \theta_\text{h},\alpha,\sigma}(Z ; I)$ be  the distribution obtained after inference~\label{method:MF-objective}. We look for
\begin{equation}
\label{method:Q_loss}
\argmax_{\theta_F, \theta_\text{u}, \theta_\text{h},\alpha,\sigma} \sum_{( Z^d,I^d)} \log Q_{\theta_F, \theta_\text{u}, \theta_\text{h},\alpha,\sigma}(Z = Z^d;I^d) \;.
\end{equation}
Since $Q_{\theta_F, \theta_\text{u}, \theta_\text{h},\alpha,\sigma}(Z = Z^d;I^d)$ is computed via a sequence of operations which are all differentiable with respect to the parameters $\theta_F, \theta_\text{u}$, and $\theta_\text{h}$, it is therefore possible to solve Eq.~\ref{method:Q_loss} by stochastic gradient descent~\cite{Domke13,Zheng15}.

\paragraph{Pre-training} 

However, it still remains difficult to optimize the whole model from scratch. We therefore pre-train our potentials separately before end-to-end fine-tuning. More precisely, the CNN $f_\text{u}$ that appears in the unary terms of Eq.~\ref{method:unary}  is trained as a standard classifier that gives the probability of presence at a given location, given the projection of the corresponding bounding-box in each camera view. For each data point, this leaves the  high-order terms for which we need to optimize
\begin{equation}
\label{method:ho_obj_1}
\sum_c \sum_{k \in \mP_c} \log(\psi^{c,k}_\text{h}(Z^d,\mF^c_k) ) \;,
\end{equation}
with respect to the parameters of the Gaussian Mixture network $\theta_\text{h},\alpha$, and $\sigma$. We use Jensen's inequality to take our generative distribution $P^g$ out of the integral in Eq.~\ref{method:ho_specific} and approximate it by random sampling procedure described in Section~\ref{sec:fullGenerative}. We rewrite the set of samples for $\vX^c_k$ from all the pixels from all the cameras from all the data-points as $S(Z^0,\dots,Z^D)$. The optimization objective of Eq.~\ref{method:ho_obj_1} can then be rewritten as
\begin{equation}
\label{method:ho_obj_2}
\sum_{\vx_s \in S(Z^0,\dots,Z^D)} \log(P^d(\vx_s | \mF^c_k,\theta_\text{h},\alpha,\sigma) ) \;,
\end{equation}
which is optimized by alternating a standard stochastic gradient descent for the $\theta_\text{h}$ parameters and a closed form batch optimization for $\alpha,\sigma$. This procedure is similar to one often used to fit a Mixture of Gaussians, except that, during the E-Step, instead of computing the class probabilities directly to increase the likelihood, we optimise the parameters of the network through gradient descent. More details are provided in the supplementary material.

This pre-training strategy creates potentials which are reasonable but not designed to be commensurate with each others. We therefore need to choose the two energy parameters scalars~$\mu_\text{u}$, and $\mu_\text{h}$, via grid-search in order to  optimize the relative weights of Unary and High-Order potentials before using the Back-Mean field method.

\subsection{Unsupervised Training}
\label{method:training:unsupervised}

In the absence of annotated training data, inter-view consistency and translation invariance still provide precious a-priori information, which can be leveraged to train our model in an unsupervised way.

Let us assume that the background-subtracting part of the network, which computes ${f_\text{b}}$, the MLP introduced in Section~\ref{sec:fullGenerative}, is reasonably initialized. In practice, it is easy to do either by training it on a segmentation dataset or by relying on simple background subtraction to compute $f_\text{b}$.  Then, starting from initial values of the parameters $\theta$, we first compute the Mean-Field approximation of $P(Z ; I_0,\theta)$, which gives us a first lower bound of the partition function. We then sample $Z$ from $Q$ and use that to train our potentials separately as if these samples were ground truth-data, using the supervised procedure of Section~\ref{method:training:supervised}. We then iterate this procedure, that is,  Mean-Field inference, sampling from $Z$, and optimizing the potentials sequentially. This can be interpreted as an Expectation-Maximization (EM)~\cite{Blei16} procedure to optimize an Expected Lower Bound (ELB) to the partition function $\mZ$ of Eq.~\ref{eq:crf}. 



%% file: implementation.tex

\section{Implementation Details}
\label{sec:implementation}

Our implementation uses a single VGGNet-16 Network with pre-trained weights. It computes features that will then be used to estimate both unary and pairwise potentials. The features map $\mF^c = \mF(I^c; \theta_{F})$ is obtained by upsampling of the convolutional layers. 

Similarly to the classification step in~\cite{Ren15}, we restrict the Region-Of-Interest pooling layer (ROI) to the features from the last convolutional layer of VGGNet. The output of the ROI is a 3x3x1024 tensor, which is flattened and input to a two layers MLP with ReLU non-linearities. In a similar way as in previous works on segmentation~\cite{Zheng15}, we use a two layers MLP to classify each hyper-column of our dense features map $\mF^c = \mF(I^c; \theta_{F})$ to produce segmentation $f_\text{b}$ and Gaussian Class $f_\text{h}$ probabilities.

We use $M = 8$ modes for Multi-Modal Gaussian distribution of Eq.~\ref{model:gaussian_mixture} for all our experiments and we have not assessed the impact of  this choice on the performance. Besides, our kernel defining the pairwise potentials of Eq.~\ref{method:binary} takes an arbitrary uniform constant value. 
%
For unsupervised training, we use a fixed number of $6$ EM iterations, which we empirically found to be enough, as illustrated in the supplementary material.

Finally, all our pipeline is implemented end-to-end using standard differentiable operations from the Theano Deep-Learning library~\cite{Theano16}. For Mean-Field inference, we use a fixed number of iterations (30) and step size (0.01). 

%% file: evaluation.tex

\section{Evaluation}
\label{sec:results}

\subsection{Datasets, Metrics, and Baselines}
\label{sec:datasets}

\noindent We introduce here the datasets we used for our experiments, the metrics we relied on to evaluate performance, and the baselines to which we compared our approach.

\paragraph{Datasets.}

\begin{itemize}

\item  \ETHZ{}. It was acquired using 7 cameras to film the dense flow of students in front of the ETHZ main building in Z\"urich for two hours. It comprises 250 annotated temporal 7-image frames in which up to 30 people can be present at a time. We used 200 of these frames for training and validation and 50 for evaluation. See the image of Fig.~\ref{fig:teaser} for a visualization.

\item \EPFL{}. The images were acquired at 25 fps on the terrace of an EPFL building in Lausanne using 4 DV cameras. The image of Fig.~\ref{fig:pom} is one of them. Up to 7 people walk around for about 3 1/2 minutes. As there are only 80 annotated frames, we used them all for evaluation purposes and relied either on  pre-trained models or unsupervised training.

\item \PETS{}. The standard {\bf PETS 2009} (PETS S2L1)  is widely used for monocular and multi-camera detection. It contains 750 annotated images and was acquired from 7 cameras. It is a simple dataset in the sense that it is not very crowded, but the calibration is inaccurate and the image quality low. 

\end{itemize}

\paragraph{Metrics.}

Recall from Section~\ref{sec:inference}, that our algorithms produces Probabilistic Occupancy Maps, such as the ones of Fig.~\ref{fig:pom}. They are probabilities of presence of people at  ground locations and are very peaky. We therefore simply label locations where the probability of presence is greater than $0.5$ as being occupied and will refer to these as {\it detections}, without any need for Non-Maximum suppression.
\comment{Should we say why they are peaky?}
We compute false positive (FP), false negative (FN) and true positives (TP) by assigning detections to ground truth using Hungarian matching. Since we operate in the ground plane, we impose that a detection can be assigned to a ground truth annotation only if they are less than a distance $r$ away. Given FP, FN and TP, we can evaluate:

\begin{itemize}

\item {\bf Multiple Object Detection Accuracy (MODA) } which we will plot as a function of $r$, and the {\bf Multiple Object Detection Precision (MODP)} ~\cite{Kasturi09}.

\item {\bf Precision-Recall}. Precision and Recall are taken to be TP/(TP + FN) and  TP/(TP+FP) respectively.
\end{itemize}
We will report MODP, Precision, and Recall for $r=0.5$, which roughly corresponds to the width of a human body. Note that these metrics are unforgiving of projection errors because we measure distances in the ground plane, which would not be the case if we evaluated overlap in the image plane  as is often done in the monocular case. Nevertheless, we believe them to be the metrics for a multi-camera system that computes the 3D location of people. 


\input{Results.tex}


\paragraph{Baselines and Variants of our Method.}

We implemented the following two baselines. 

\begin{itemize}

\item \POM{}.  The multi-camera detector~\cite{Fleuret08a} described in Section~\ref{related:subsec:multicam} takes background subtraction images as its input. In its original implementation, they were obtained using traditional algorithms~\cite{Ziliani99,Oliver00}. For a fair comparison reflecting the progress that has occurred since then, we use the same CNN-based segmentor as the one use to segment the background, that is $f_\text{b}(\mF^c_k;\theta_{b})_0$ from Eq.~\ref{model:gaussian_mixture}.

\item  \CNN{}.  The recent work of~\cite{Xu16} proposes a MCMT tracking framework that relies on a powerful CNN for detection purposes~\cite{Ren15}, as discussed in Section~\ref{related:subsec:multicam}.   Since the code of~\cite{Xu16}  is not publicly available, we reimplemented their detection methodology as faithfully as possible but {\it without} the tracking component for a fair comparison with our approach that operates on images acquired at the same time. Specifically, we run the 2D detector~\cite{Ren15} on each image. We then project the bottom of the 2D bounding box onto the ground reference frame as in~\cite{Xu16} to get 3D ground coordinates. Finally, we cluster all the detections from all the cameras using 3D proximity to produce the final set of detections. 

\end{itemize}

To gauge the influence of the different components or our approach, we compared these baselines  against the following variants of our method. 

\begin{itemize}

\item \OURSWFT{}. Our method with all three terms in the CRF model turned on, as described in Section~\ref{sec:completeCRF}, and fine tuned end-to-end through back Mean-Field, as described in Section~\ref{method:training:supervised}.

\item \OURSNFT{}. \OURSWFT{} without the final fine-tuning. 

\item \OURSUSP{}. Same as \OURSNFT{} but the training is done without  ground truth annotations, as described in Section~\ref{method:training:unsupervised}. 

\item \OURSSHO{} : We replace the full High-Order term of Section~\ref{sec:hoCRF}  with the simplified one that approximates the one of~\cite{Fleuret08a}, as described at the beginning of that section.

\item \OURSNHO{}.  We remove the High-Order term of Section~\ref{sec:hoCRF}  altogether.

\end{itemize}

\subsection{Results}
\label{sec:results}

We report our results on our three test datasets in Fig.~\ref{fig:EpflEthz}.

\vspace{-0.4cm}
\paragraph{\ETHZ{}.}

\OURSWFT{} and  \OURSNFT{} clearly dominate the \CNN{} and \POM{} baselines, with \OURSWFT{} slightly outperforming \OURSNFT{} because of the fine-tuning. Simplifying the high-order term, as in \OURSSHO{}, degrades performance and removing it, as in  \OURSNHO{}, degrades it even more.  The methods discussed above rely on supervised training, whereas \OURSUSP{} does not but still outperforms the baselines.

\vspace{-0.4cm}
\paragraph{\EPFL{}.}

Because the images have different statistics than those of \ETHZ{}, the unary terms as well as the people detector \CNN{} relies on are affected. And since there is no annotated data for retraining, as discussed above, the performance of \OURSNHO{} and \CNN{} drop very significantly with respect to those obtained on \ETHZ{}. By contrast, the high order terms are immune to this, and both \OURSNFT{} and  \OURSUSP{} hold their performances. 


\vspace{-0.4cm}
\paragraph{\PETS{}.}

The ranking of the methods is the same as before except for the fact that  \OURSSHO{} does as well as \OURSNFT{}. This is because the~\PETS{} dataset is poorly calibrated, which results in inaccurate estimates of the displacement vectors in the generative model of Section~\ref{sec:fullGenerative}. As a result, it does not deliver much of a performance boost and we therefore did not find it meaningful to report results for unsupervised training and fine-tuning of these High-Order potentials.

\paragraph{From Detections to Trajectories.}

Since our method produces a Probability Occupancy Map for every temporal frame in our image sequences, we can take advantage of a simple-flow based method~\cite{Berclaz11} to enforce temporal consistency and produce complete trajectories. As shown in Fig.~\ref{tab:ksp} this leads to further improvements for all three datasets. 


\begin{figure}[ht!]
\vspace{-0.1cm}
\centering
\begin{tabular}{|l|c|c|c|}
\hline
\textbf{Method} & \ETHZ{} & \EPFL{} & \PETS{}   \\ \hline
Ours            &   74.1\%  & 68.2\%  & 79.8\% \\ \hline
Ours + \cite{Berclaz11}    &  75.2\%   & 76.9\%  & 83.4\% \\ \hline
\end{tabular}
\vspace{-0.2cm}
 \caption{MODA scores for $r=0.5$ before and after enforcing temporal consistency. }
 \label{tab:ksp}
\end{figure}

%% file: Results.tex

\begin{figure*}[ht!]
\begin{center}
\begin{tabular}{ccc}
\vspace{-0.2cm}
ETHZ & EPFL & PETS \\
\vspace{-0.0cm}
\hspace{-0.8cm}\includegraphics[width=0.31\textwidth]{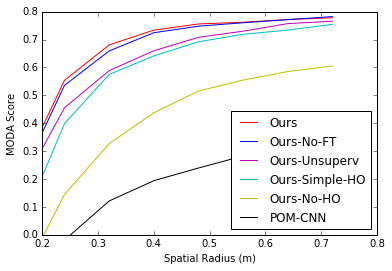} &
\includegraphics[width=0.31\textwidth]{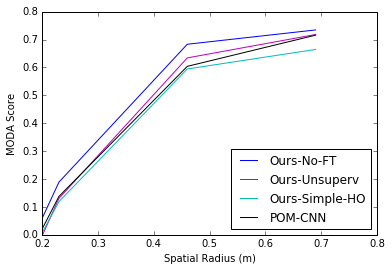} &
\includegraphics[width=0.31\textwidth]{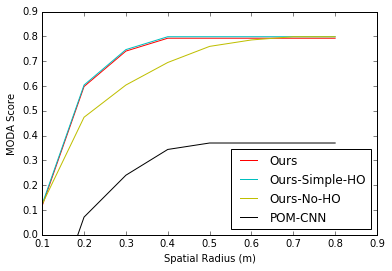} \\
\end{tabular}
\\[-0.2cm]
\begin{tabular}{|c|c|c|c|c|c|c|}
\hline
\textbf{}       & \multicolumn{2}{c|}{\ETHZ{}}                  & \multicolumn{2}{c|}{\EPFL{}} & \multicolumn{2}{c|}{\PETS{}} \\ \hline
\textbf{Method} & \textbf{Precision / Recall} & \textbf{MODP} & \textbf{Precision / Recall} & \textbf{MODP}   & \textbf{Precision / Recall} & \textbf{MODP}   \\ \hline
\OURSWFT{}    & \textbf{95 / 80\%}       & \textbf{53.8\%}          &         -                        &      -                             &         -                        &         -      \\ \hline
\OURSNFT{}    & 93 / 80\%                     & 53.4\%                       &   \textbf{88 / 82\%} &    \textbf{48.3\%}      & \textbf{93 / 87\%}    &    \textbf{60.4\%}\\ \hline
\OURSUSP{}    & 86 / 80\%                    & 49.8\%                       &   80 / 85\%               &    47.5\%                    &       -                           &       -        \\ \hline
\OURSSHO{}    & 87 / 70\%                    & 47.5\%                       &   85 / 75\%               &    43.2\%                    & 93 / 87\%                 &    60.4\%  \\ \hline
\OURSNHO{}    & 84 / 55\%                    & 34.4\%                       &  37 / 68\%                &    23.3\%                    & 93 / 81\%                 &    55.2\% \\ \hline
\POM{}        & 75 / 55\%                            & 30.5\%                       &  80 / 78\%                &   45.9\%                     & 90 / 86\%                 &    42.9\% \\ \hline
\CNN{}        & 68 / 43\%                            & 18.4\%                       &   39 / 50\%               &    21.6\%                    & 50 / 63\%                &    27.6\%  \\ \hline
\end{tabular}
%
%
\end{center}
\vspace{-0.2cm}
   \caption{Results on our three test datasets. {\bf Top row.} MODA scores for the different methods as function of the radius $r$ used to compute it, as discussed in Section~\ref{sec:datasets}.  {\bf Bottom row.}  Precision/Recall and MODP for the different methods for $r=0.5$. Some of the values are absent either due to the bad calibration of the data-set, or missing ground-truth, as explained in Sections~\ref{sec:datasets} and~\ref{sec:results}. The numbers we report for the \CNN{} baseline are much lower than those reported in~\cite{Xu16} for the method that inspired it, in large part because we evaluate our metrics in the ground plane instead of the image plane and because~\cite{Xu16} uses a temporal consistency to improve detections.}
\label{fig:EpflEthz}
\vspace{-0.3cm}
\end{figure*}

%% file: conclusion.tex

\section{Discussion}
\blfootnote{ This work was supported in 
part by the Swiss National Science Foundation, under the grant CRSII2-147693 ``Tracking in the
Wild''.}

We introduced a new CNN/CRF pipeline that outperforms the state-of-the art for multi-camera people localization in crowded scenes. It handles occlusion while taking full advantage of the power of a modern CNN and can be trained either in a supervised or unsupervised manner.

A limitation, however, is that the CNN used to compute our unary potentials still operates in each image independently as opposed to pooling very  early the information from multiple images and then leveraging the expected appearance consistency across views. In future work, we will therefore investigate training such a CNN for people detection on multiple images simultaneously, jointly with our CRF. 



